\documentclass[10pt,twocolumn,letterpaper]{article}

\usepackage{cvpr}              

\usepackage{graphicx}
\usepackage{amsmath}
\usepackage{amssymb}
\usepackage{booktabs}
\usepackage{xcolor}
\usepackage{textcomp}
\usepackage{gensymb}

\usepackage[pagebackref,breaklinks,colorlinks]{hyperref}

\usepackage[capitalize]{cleveref}
\crefname{section}{Sec.}{Secs.}
\Crefname{section}{Section}{Sections}
\Crefname{table}{Table}{Tables}
\crefname{table}{Tab.}{Tabs.}

\makeatletter
\newcommand{\printfnsymbol}[1]{%
        \textsuperscript{\@fnsymbol{#1}}%
}
\makeatother

\def\cvprPaperID{23} 
\def\confName{CVPR}
\def\confYear{2023}

\begin{document}

\title{Robot Goes Fishing: Rapid, High-Resolution Biological Hotspot Mapping in Coral Reefs with Vision-Guided Autonomous Underwater Vehicles}
\author{
$\text{Daniel Yang}^{1,2*} \qquad$
$\text{Levi Cai}^{1,2*} \qquad$
$\text{Stewart Jamieson}^{1,2} \qquad$
$\text{Yogesh Girdhar}^{2} \qquad$
\\
$^1\text{Massachusetts Institute of Technology} \qquad ^2\text{Woods Hole Oceanographic Institution}$
\\
${\small\texttt{\{dxyang,cail,sjamieso\}@mit.edu, yogi@whoi.edu}}$
}


\twocolumn[{%
\renewcommand\twocolumn[1][]{#1}%
\maketitle
\begin{center}
    \newcommand{\teaserwidth}{\textwidth}
    \vspace{-0.35in}
    \centerline{
       \includegraphics[width=\teaserwidth,clip]{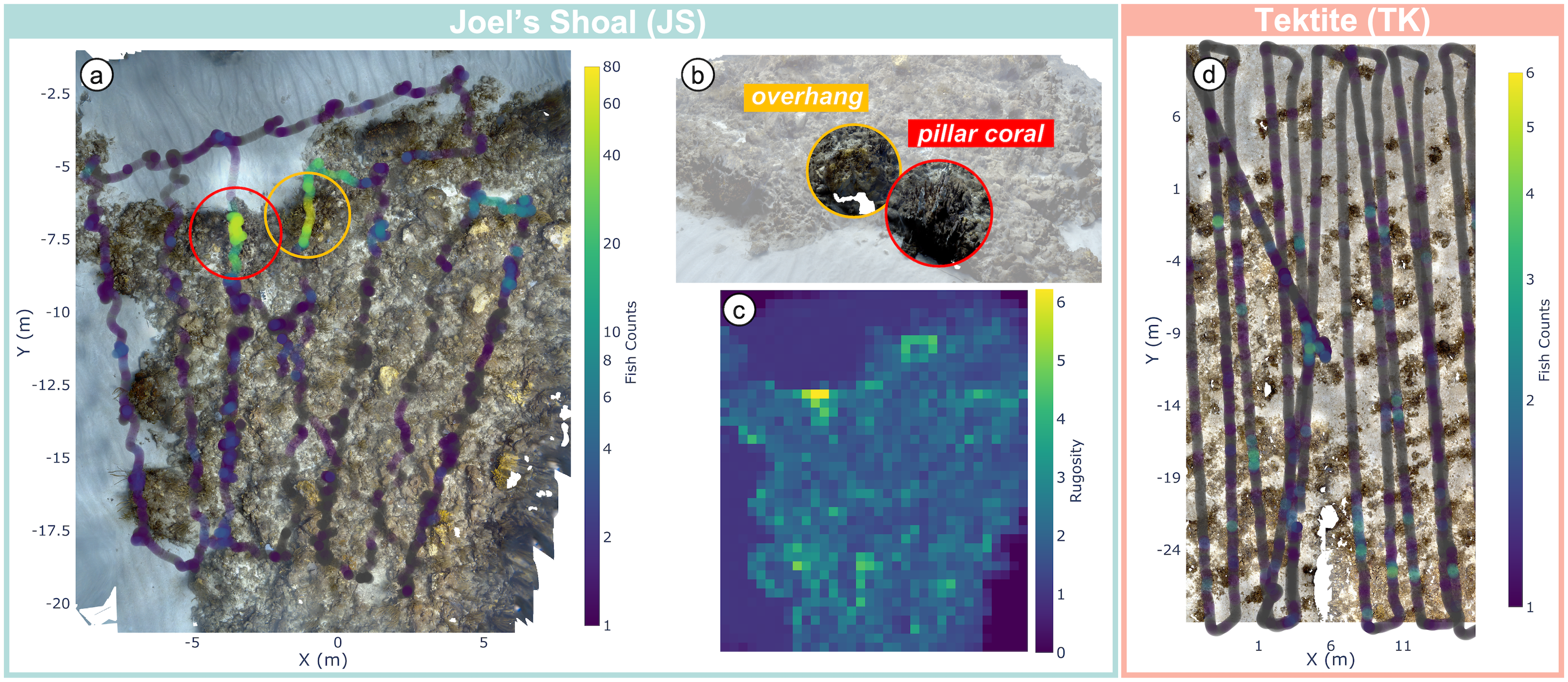}
    }
    \captionof{figure}{Hotspot detection at two unique coral reef sites in the US Virgin Islands --- Joel's Shoal (JS) and Tektite (TK). From visual surveys collected by an AUV, we generate hotspot maps of fish counts along the robot trajectory and plotted with a log-colorscale (a/d). Our fish detector is fine-tuned on labelled images from JS, but not from TK. We can see distinct hotspots at both reefs, correlated with distinct reef structure visible in our 3D reconstruction, an overhanging structure and a pillar coral (b), as well as with rugosity (c), which is a measure of topographic complexity used by the biological community as a proxy for biodiversity and abundance. Our system enables rapid deployment of an AUV over a predefined but unknown area to identify biologically interesting hotspots.}
    \vspace{-0.05in}
    \label{fig:teaser}
\end{center}
}]


{\let\thefootnote\relax\footnote{\vspace{-0.1in}
\tiny \par *Authors contributed equally to this work.
\par
\it{This material is based upon work supported by the National Science Foundation under Grant No.. 2133029.}}}


\begin{abstract}
\vspace{-0.1in}  
Coral reefs are fast-changing and complex ecosystems that are crucial to monitor and study. Biological hotspot detection can help coral reef managers prioritize limited resources for monitoring and intervention tasks. Here, we explore the use of autonomous underwater vehicles (AUVs) with cameras, coupled with visual detectors and photogrammetry, to map and identify these hotspots. This approach can provide high spatial resolution information in fast feedback cycles. To the best of our knowledge, we present one of the first attempts at using an AUV to gather visually-observed, fine-grain biological hotspot maps in concert with topography of a coral reefs. Our hotspot maps correlate with rugosity, an established proxy metric for coral reef biodiversity and abundance, as well as with our visual inspections of the 3D reconstruction. We also investigate issues of scaling this approach when applied to new reefs by using these visual detectors pre-trained on large public datasets.
\end{abstract}


\vspace{-0.2in}  

\section{Introduction}
\label{sec:intro}


Coral reefs are essential ecosystems for their biodiversity \cite{plaisance2011diversity} and are identified as crucial environments to protect and monitor \cite{jackson2014status, towle_national_2021}. However, reefs are highly dynamic environments with complex spatial structures and their ecosystem can change rapidly due to weather events, anthropogenic causes, or diseases \cite{harvell1999emerging}. This necessitates the development of methods to quickly and continually assess ecosystem health at fine granularity.

Current techniques for assessing and monitoring coral reef conditions in high resolution rely heavily on human SCUBA divers, a time and labor intensive process requiring specialized training and potentially exposing humans to physiologically dangerous conditions \cite{loerzel2017scuba}. Processing of data products gathered during these dives (e.g., video footage) may also be time and labor intensive, requiring dense annotation with specialized knowledge (e.g., how to identify different animal species, coral reef diseases). 

Loiseau et al. \cite{loiseau_indices_2015} identify that most metrics used to assess and respond to coral reef health rely strictly on number of species present in large portions of reefs. They propose that this single metric is insufficient for many conservation or monitoring tasks. The NOAA 2021 National Coral Reef Monitoring Plan and Shantz et al. \cite{shantz_fish-derived_2015, towle_national_2021} suggest that benthic animals tend to create recurring community structures around specific coral heads and form biodiverse ``hotspots'' that are critical to monitoring efforts. To find these coral heads, scientists quantify complex topographies by mapping rugosity, the ratio between true and geometric surface area change for a region, which is correlated with species diversity and abundance in coral reefs \cite{fuad_coral_nodate,santoso2022influence, yanovski_structural_2017}. Current state-of-the-art methods in mapping reef fish use acoustics \cite{Costa_fishacoustics_2014}, which can be used at large spatial scales (100 km$^2$) but at a very coarse resolution of about 100 m$^2$. However, small hotspots (1 m$^2$) can be crucial for the functioning of many reef ecosystems. Meanwhile, Francisco et al. \cite{francisco_high-resolution_2020} note that AUVs with cameras and deep learning can be used to provide fast, high-resolution mapping and tracking capabilities in benthic environments. However, their focus was on single or small groups of organisms. 

Here, we use AUVs and vision to (1) map benthic areas, (2) identify significant coral structures, and (3) find fine-grain hotspot communities for scientists and conservationists to prioritize during monitoring. Broadly, we propose to use AUVs to survey pre-defined, though not necessarily pre-mapped, benthic regions using a downward-facing camera. The collected imagery can be used in structure-from-motion pipelines to generate three-dimensional maps of the environment and provide fine-grain localization details. Those same images can be processed by visual object detection models, such as YOLO \cite{redmon_you_2016} or RetinaNet \cite{lin2017focal}, trained to detect marine organisms, and provide organism counts at a frame-level. By co-locating these detections over maps, we can determine hotspots over coral reefs at resolutions as fine as individual coral heads. This approach allows us to gather fast, high-resolution data. While similar approaches have been explored with drones in other conservation domains \cite{van2015nature}, less has been done underwater.


\section{Technical Approach}
\label{sec:approach}
Our approach enables rapidly detecting biological hotspots in novel environments with minimal human intervention. We first use an AUV to collect a visual survey of the unexplored environment, with no \textit{a prori} knowledge other than the rough spatial bounds. This visual survey is processed to (1) generate a 3D reconstruction of the environment and (2) estimate fish counts at every frame. By localizing these fish counts onto precise frame locations within the 3D environment, we then have a visual map of fish abundance across the full area of interest, allowing correlation of biological activity with local reef structures. 

\subsection{Mapping coral reefs with an AUV}
\label{sec:methods-mapping}

We utilize the 
Curious Underwater Robot for Ecosystem Exploration (CUREE) \cite{girdharcuree2023} 
AUV platform to conduct our experiments but note that our approach is system agnostic, assuming only visual sensing and state estimation capabilities. The AUV can perform low-altitude benthic visual surveys at relatively high frame rates and resolutions to ensure overlap between frames and sufficient resolution to capture fine-grain coral reef features. We generate fixed ``lawnmower'' trajectories for the AUV to follow that achieve dense coverage of a coral reef site at a constant altitude of 2m while collecting visual survey data using the downward-facing camera running at 4K resolution at 6 FPS with a FOV of 120\textdegree$\times$58\textdegree. Currently, the AUV uses an on-board doppler velocity log (DVL) to maintain altitude, though visual SLAM could be used for additional control and enable even lower-altitude surveys.

Using the high-resolution, downward facing data collected, we can use off-the-shelf photogrammetry software, Metashape~\cite{metashape}, to generate both detailed 3D reconstructions as well as 2D top-down orthomosaic images of the environments explored by the AUV. This reconstruction allows us to also estimate the $SE(3)$ pose of the camera at every image relative to a constant world frame, thus allowing us to align information extracted from images (e.g., fish counts) with precise spatial locations on the reef site. 

We compute the surface area of the 3D mesh over each 0.5m$\times$0.5m grid cell in the XY-plane and divide by the cell area (0.25 m$^2$) to compute \textit{rugosity}.  Notably, this approach to rugosity estimation is minimally invasive as it does not require any close physical interaction with reef structures.

\subsection{Hotspot identification via fish detection}
\label{sec:methods-yolo2hotspot}

Using the same imagery collected and localized in \Cref{sec:methods-mapping}, we estimate fish counts, or abundance, in each frame (note that averaging fish abundance across frames is difficult since individual tracking across frames needs to be accounted for), which correlates to immediate fish abundance at a particular location as given by the generated map. For this work, we focus only to detect objects of a single class, \textit{fish}, as opposed to fine-grain species classification.

To provide fast, dense, and automated estimation of fish counts, we rely on a state-of-the-art visual detector based on deep learning methods. In this work, we use the YOLOv5 detector \cite{jocher_yolo} with an eye towards real time applications on AUVs in the field, though any visual detector could be used. Naively, a highly accurate detector is sufficient to generate the desired abundance estimates and provide a hotspot map as shown in \Cref{fig:teaser}. However, deep neural network-based detectors introduce a variety of issues that must be addressed before results from these types of approaches can be used to inform coral reef managers, principally in uncertainty and error quantification as discussed in \Cref{sec:limitations}.

To estimate biological hotspots within surveyed environments, we directly apply our model over the downward facing images from the visual survey. Enabled by the 3D reconstructions in Section \ref{sec:methods-mapping}, we can precisely assign each geospatial location visited along the AUV trajectory with an estimated fish count, as measured by the number of detections in a frame. Our use of a downward facing camera, where the visual frustum is relatively concentrated over an area, enables us to directly assign fish counts to camera poses, whereas the use of a forward facing camera where the horizon is visible would complicate such an assumption. 

Ideally, these measurements could be used to directly assign fish abundance (or population estimate) to specific regions. However, as further discussed in \Cref{sec:limitations}, these models are prone to false-positive detections, which lead to overcounting, and false-negative detections, or undercounting. These issues can be exacerbated when deploying on out-of-sample data \cite{koh2021wilds}, such as on new reefs, or even the same reef at a different time or under different turbidity conditions. Thus, while further work in uncertainty quantification is needed to estimate absolute abundance, our work focuses on \emph{relative} abundance within a visual survey, which can be used to determine hotspots.

We investigate two strategies for deploying these detectors for hotspot mapping: (1) \textit{interpolation} of \textit{in-sample} survey data and (2) \textit{extrapolation} to \textit{out-of-sample} coral reef surveys. For interpolation, we manually annotate one frame every 20 seconds of video as well as the frame one second before and after each of our 20 second intervals. This process is motivated by trading off between covering different scales of variance within our dataset --- the coarse 20 second intervals allow us to see different species and conditions across a visual survey while the one second intervals capture the same fish detections under different poses. These intervals are also chosen under the constraints of limited human annotation time. We can then train a detector, in this case we use YOLOv5m \cite{redmon_you_2016} with 1280p resolution, and use it to generate dense estimates in all the intermediate unlabelled frames, hence in-sample interpolation.

However, for these detectors to be useful and drastically reduce annotation effort, they must detect fish with similar accuracy on other reefs, or at different times in various conditions. We thus provide an initial investigation by deploying the same detector used for the in-sample study, and apply it without modification to a survey on another coral reef. We minimize changes in the environment by selecting a reef that is nearby to the original, and survey at a similar time of day and weather conditions, only one day apart. 

Next, we also investigate using a fully out-of-sample detector, which we call the \textit{MegaFishDetector} (MFD), similar to \cite{beery2019efficient}, but trained using several publicly available fish detection datasets using the same YOLOv5m at 1280p network. This dataset consists of roughly 210K training images and 36K validation images, assembled together from the following datasets: AIMS OzFish \cite{marrable_accelerating_2022}, FathomNet (only the \textit{gnathostomata} class) \cite{katija_fathomnet_2022}, VIAME FishTrack22 (subsampled by 50\%) \cite{dawkins2022fishtrack}, NOAA Puget Sound Nearshore Fish \cite{ferriss_characterizing_2021}, NOAA Labeled Fishes In the Wild \cite{cutter2015automated}, and DeepFish (Segmentation only) \cite{saleh_realistic_2020}. Finally, we compare the ability of using MFD alone or as a pre-trained network that can be used to fine-tune the in-sample network. MFD code and models are available at \url{https://github.com/warplab/megafishdetector}.


\section{Results}
\label{sec:results}

\begin{table}[]
\resizebox{\columnwidth}{!}{%
\begin{tabular}{lllllll}
\hline
Site         & Size      & Duration & Date (d/m/y) & Time    & \# imgs & \# label imgs \\ \hline
Joel's Shoal (JS) & 12m x 12m & 36min    & 03/11/2022   & 11:42AM & 13236   & 312           \\
Tektite (TK)     & 40m x 20m & 37min    & 04/11/2022   & 12:59PM & 13666   & 113           \\ \hline
\end{tabular}%
}
\caption{Information about the sites, surveys performed (4K resolution, 6fps imagery), and datasets created as part of our field trials in St. John, U.S. Virgin Islands, USA in November 2022.}
\label{tab:sites}
\end{table}

\begin{table}[]
\resizebox{\columnwidth}{!}{%
\begin{tabular}{llllll}
\toprule
Model   & Training data & Evaluation data & In sample? & mAP50          & mAP95          \\ \midrule
MFD     & MFD           & JS              & No         & 0.012          & 0.00412        \\
JS      & JS            & JS              & Yes        & 0.895          & 0.58           \\
$\text{JS}_{\text{MFD}}$ & JS            & JS              & Yes        & \textbf{0.914} & \textbf{0.782} \\
MFD     & MFD           & TK              & No         & 0.0243         & 0.00726        \\
JS      & JS            & TK              & No         & 0.127          & 0.06           \\
$\text{JS}_{\text{MFD}}$ & JS            & TK              & No         & \textbf{0.225} & \textbf{0.115} \\ \bottomrule
\end{tabular}%
}
\caption{Quantitative analysis on the performance of the Yolov5m model, at 1280p, with different training configurations. MFD refers to the MegaFishDetector dataset, JS is the Joel's Shoal dataset, and JS$_{MFD}$ is pretrained on MFD and finetuned on JS. We see that MFD alone is insufficient, and JS alone is still not robust even to small changes in site context. JS$_{MFD}$ is slightly more robust to out of sample data, but still more work needs to be done.}
\label{tab:quant-results}
\end{table}

\begin{figure}
    \centering
    \includegraphics[width=\linewidth]{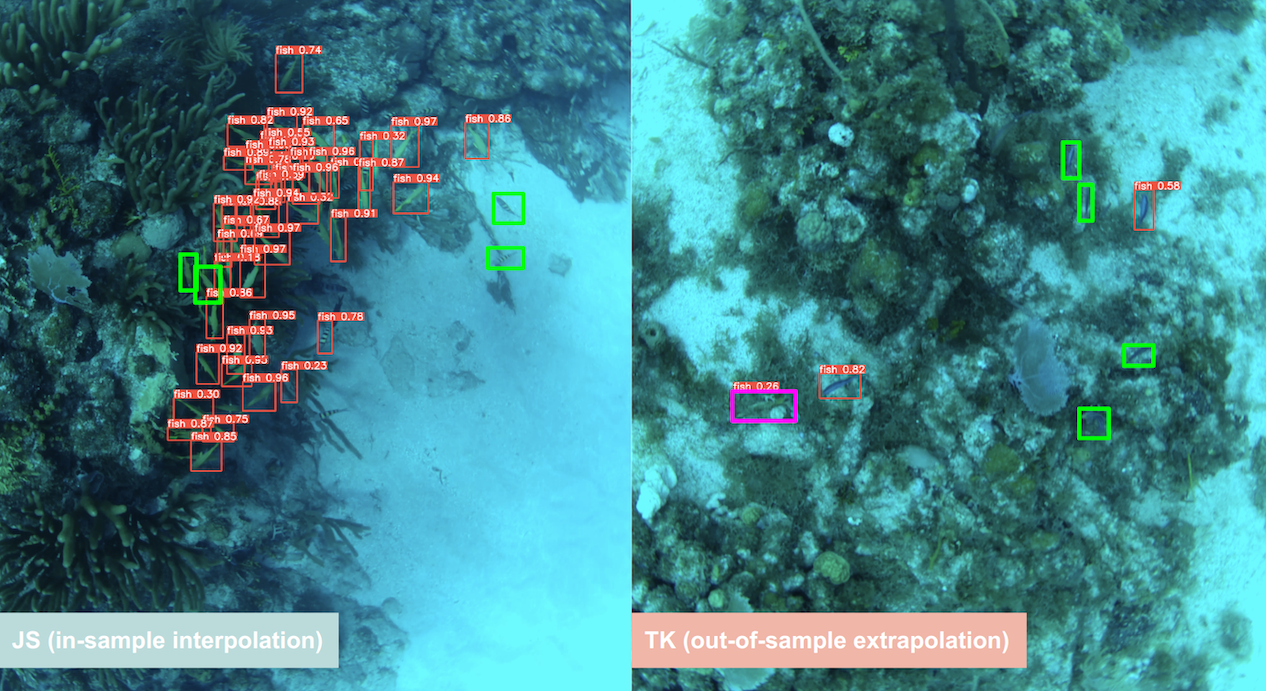}
    \caption{Qualitative examples of YOLO output on in-sample and out-of-sample data. In red are predictions with confidence numbers, green are false-negatives (undercounted), magenta are false-positives (overcounted).}
    \label{fig:qual_yolo}
\end{figure}

\subsection{Field trials in the US Virgin Islands}

We collected surveys at two unique coral reefs in the US Virgin Islands in November 2022 --- Joel's Shoal (JS) and Tektite (TK)--- as described in \Cref{tab:sites}. For in-sample experiments, our fish detectors are trained only on labelled images from the Joel's Shoal survey. For the 36 min survey, our annotation process yields a dataset of 312 labelled frames that we divide into a 75\% train and 25\% validation dataset. Due to the low data count, we did not include a test dataset for the JS survey, for in-sample verification, we re-run the detector on all images (training and validation). Using this data and MFD data, we train three different networks, the results are shown in \Cref{tab:quant-results}. The first is only using MFD data, which trained for 50 epochs. The second is using only JS data, which trained for 232 epochs (until mAP no longer improved). The third is training a network using only the JS data, but is finetuned for 600 epochs and initialized from the MFD detector we already trained. In all cases, our networks initialized by a network pre-trained on MS-COCO \cite{cocodataset} and are based on the YOLOv5m model and use 1280p resolution. For out-of-sample experiments, we evaluate on data from Tektite, a different coral site which we did not include in training data. TK is located roughly 1.5 km away from JS, with the survey almost 24 hrs apart, and so are in similar conditions with similar biodiversity.

From the results in \Cref{tab:quant-results}, we see that MFD alone cannot provide reasonable abundance metrics, with an mAP50 of 0.012. While in-sample networks on the JS dataset achieve an mAP50 of 0.895, reasonable for interpolation, performance is drastically reduced when used out-of-sample, dropping to an mAP50 of 0.127 on the TK dataset. Finally, we see that MFD pre-training and fine-tuning with some context-specific data of JS, resulting in model JS$_{MFD}$, allow us to achieve an mAP50 of 0.225 on the out-of-sample TK data, but is still not sufficient for wide-spread application.

In \Cref{fig:teaser}, we show our a log-count map of fish abundance overlaid on an orthomosaic of the whole site generated by the JS survey. To our knowledge, this is the first fine resolution (\textless 1 m error) map of marine life abundance collected by an AUV. Visually, we see two distinct hotspots along the AUV trajectory at approximately (-3.5, -7.0) m and (-1.0, -6.5) m which correspond to a pillar coral, \textit{dendrogyra}, and a soft coral above an overhung structure, as shown in the 3D reconstructions. High diversity and abundance of marine life is known to congregate in regions of topographic complexity, like pillar corals and overhanging areas, as they can provide shelter  \cite{mazzuco2020substrate, menard2012shelters}. Likewise, we see a lower amount of detections over sandy, flatter regions off the coral reef shelf. This correlation between topographic complexity and fish abundance is highlighted when comparing our hotspot map with the rugosity plot in \Cref{fig:teaser}, where high rugosity correlates with high abundance. We also see that rugosity is not strictly predictive of abundance, and thus the use of visual detector is still necessitated.



In contrast to the in-sample training domain data processed to generate the previous fish abundance map, in \Cref{fig:teaser}, we show results on a different visual survey conducted at TK. While JS consists of one large island of coral reefs bordered by a sand, TK is much more sporadic, containing clusters of reefs over a larger sandy region. Our plot indicates correlation between these coral regions and fish abundance, but there is no distinct hotspot as in JS. Note that the survey area at TK is almost six times larger, highlighting that an AUV can rapidly cover large regions with minimal human intervention. 



\section{Conclusion and Discussion}
\label{sec:conclusion}

In this work, we proposed and demonstrated a potential pipeline for using vision-guided AUVs to identify biological hotspots in coral reef ecosystems. In a single pass, an AUV can provide a high-resolution 3D map, map rugosity, and estimate fish abundance using visual detectors, all using the same imagery. These localizations can be provided at high-resolution and performed quickly which in turn help coral reef managers monitor and respond to events across reefs with high precision. We also trained generic fish detectors from large-scale data to investigate our ability to generalize to new coral reefs. 

\label{sec:limitations}
While our approach provides immediate value for ecosystem monitoring, there are many limitations to study in future work. Better detectors are required to improve performance in out-of-sample situations, so we can deploy these systems on new reefs, or even if conditions change in an previously surveyed reef. MegaFishDetector is comprised largely of \textit{baited} and forward-looking imagery, in non-coral reef environments, which does not correlate well with our visual environment. Furthermore, novel conditions are unlikely to be avoided, so better methods for determining out-of-sample confidence, and propagating those uncertainties to abundance metrics will help ensure accuracy. 


Additionally, AUVs can impact the behavior of ecosystem inhabitants which may skew detections. To ensure we do not damage these sensitive ecosystems, our AUV maintains a minimal altitude of 2m, which with our high FOV, means that many fish are only a few pixels wide. Finally, many fish naively resemble the corals, and color correction or accounting for motion may enhance performance.


{\small
\bibliographystyle{ieee_fullname}
\bibliography{main,veevee}
}

\end{document}